# Real-World Semantic Grasp Detection Based on Attention Mechanism*

Mingshuai Dong, Shimin Wei, Jianqin Yin, Xiuli Yu*

*Abstract*—Recognizing the category of the object and using the features of the object itself to predict grasp configuration is of great significance to improve the accuracy of the grasp detection model and expand its application. Researchers have been trying to combine these capabilities in an end-to-end network to grasping specific objects in a cluttered scene efficiently. In this paper, we propose an end-to-end semantic grasp detection model, which can accomplish both semantic recognition and grasp detection. And we also design a target feature attention mechanism to guide the model focus on the features of target object ontology for grasp prediction according to the semantic information. This method effectively reduces the background features that are weakly correlated to the target object, thus making the features more unique and guaranteeing the accuracy and efficiency of grasp detection. Experimental results show that the proposed method can achieve 98.38% accuracy in Cornell Grasp Dataset. Furthermore, our results on complex multi-object scenarios or more rigorous evaluation metrics show the domain adaptability of our method over the state-of-the-art.

*Keywords*：**Grasp detection, attention mechanism, semantic segmentation**

## I. INTRODUCTION

Semantic grasp detection is the key ability of the robot to interact with the environment. Robotic semantic grasping aims to find the target object in a cluttered environment and predict the grasp configuration according to its features. Traditionally, robotic grasp detection is commonly achieved by traversing the features of the whole image [1-8] without distinguishing the target features and background features. Therefore, humans can easily grasping a specified object in a cluttering scene, but it is far from an easy problem for a robot to solve.

Recognition and location of the target and selecting the features of the target area are the key steps to realizing the semantic grasping of the robot. With the development of deep learning technology, object detection models based on deep learning have achieved satisfactory results [9]. On this background, [10] [11] [12] proposed a multi-task model that simultaneously implements object detection and grasp detection. These multi-task models give the grasp position category information according to the relationship between the predicted grasp position and the target position to realize the grasping of a specific object. Yet, its grasp detection depends on the features of the whole image. In most cases, it is feasible to guide the model to determine the target feature area and predict a viable gripper position using the features within this small range. Because the working scene of the robot is chaotic and changeable, and the model is susceptible to the influence of ambiguity features in the clutter background so as to predict the wrong grasp configuration. As shown in Fig. 1(b) *Feature Map*, there are many features similar to the target object in the image, which will lead the model to provide wrong prediction results. Therefore, the features selection and location matter much more to robotic grasp detection.

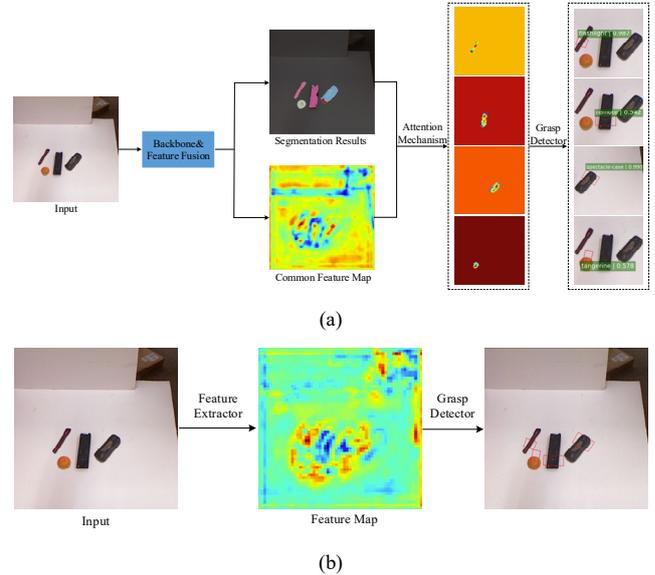

Figure 1. (a): The realization process of semantic grasp detection based on attention mechanism. Input RGB image through feature extraction and feature fusion module to generate common feature map. The feature attention mechanism guides the model to use the features of the object itself to predict the grasp configuration. (b): Previous grasp detection. The previous grasp detection approach directly uses all the features of the feature map to predict the grasp configuration.

Recently, [13] proposed identifying the target object's category and ROI (Region of Interest) region through the object detection method and only using the features of the ROI region to predict the grasp pose. This method reduces the area of grasp detection, and the model pays more attention to the features of the target object. However, due to the axial symmetry of ROI, there will still be background features and local features of other objects in the ROI region, thus affecting the model's performance. To solve the above problems, some learning-based algorithms further reduce the background features that contribute little to the detection by using semantic segmentation [14]. However, the redundant feature extraction structure of this method requires huge computing power and time.

In this paper, we discard redundant structures in the network and use an end-to-end model to generate grasp predictions for specified objects with quality scores. Our model implements semantic segmentation and grasp detection simultaneously via sharing one single backbone. In order to improve the representation ability of feature maps, we adopt a feature fusion network that merges low-level features with high-level features to generate feature maps containing rich semantic and structural information. Moreover, we design a feature attention mechanism to guide the model to use the features of the object itself to predict the grasp configuration according to the result of semantic segmentation. Compared with the existing methods, our model is a dense end-to-end model, which predicts the grasp configuration according to the ontology features of the specified target object. Our approach is closer to the human object-oriented grasp detection method and has higher accuracy than the ROI-based [13] method and state-of-the-art method [2].

The main contributions of this paper are concluded as follows:

In this paper, we propose a dense end-to-end grasp detection model focusing on the object itself features, which is 2.87% higher in accuracy than the state-of-the-art grasp detection algorithm. And even under the harsh experimental parameters, the model still achieves reasonable results.

Pointing at the characteristics of robot semantic grasping, we propose a target feature attention mechanism. This mechanism enables our model to be extended to multi-object scenes to grasp a specific object. And we design ablation experiments to prove the contribution of this mechanism to grasp detection.

The rest of this paper is organized as follows: We first review the background and related works of our approach in Sec II. Then in Sec III, we introduce grasp representation in RGB image and prediction method of our model in the multi-object scenes. Sec IV describes the details of our model and the realization principle of the feature attention mechanism. In Sec V, we give the dataset used for model training and the metrics of the model. In Sec VI, we design ablation experiments, comparative experiments, and physical experiments to demonstrate our contributions and the effectiveness of the proposed model. Finally, in Sec VII, we discuss the conclusions of this paper and future work.

## II. RELATED WORK

At present, researchers mainly adopt the analytic approach [15], model-based approach [16], and learning-based approach for grasp detection. The analytic approach [17], [18] refers to finding the region in the point cloud that satisfies the specific predefined grasping requirements. This method can efficiently predict reliable robot grasping posture but requires a lot of hand-engineered features. Model-based approach [19], [20] takes the object's 3D model as input to predict the grasping pose. People predefined special grasping posture according to different object shapes. Therefore, such approaches have shortcomings in predicting the novel object's grasping pose. The learning-based grasp detection method becomes possible with the popularization of deep learning technology and the pretraining model. And features extracted through deep learning models proved to be better than hand-engineered ones. Hence, the learning-based approach has attracted great attention from researchers. This section will focus on the learning-based approach and the direction that researchers have taken to improve it.

[21] first proposed to represent the grasping position of the target object in RGB or RGB-D image with an orientation rectangle. This method transforms the complex grasp configuration detection into a problem similar to object detection and lays a foundation for later grasp detection research. Based on this method, [1] sped up the grasp detection by local constrained prediction mechanism to divide the image into multiple regions and then use the direct regression method to predict the possible grasping positions in each region. This method addresses the issue that the direct regression algorithm can only predict one grasping position in the input image and improves the accuracy of the grasp detection model. [4] further improves the grasping accuracy of the model by using ResNet-50 as the backbone network to directly regression the grasp configuration in RGB-D images. This method demonstrates that the deeper the network, the better the performance of the grasp detection model. The above methods are based on the scene features to direct regression the grasp configuration. However, due to the complexity of the robot working scene, the training of the direct regression algorithm is complicated and requires a quantity of time and computing resources.

In order to alleviate the problems of training difficulties and limited accuracy of the direct regression model, researchers proposed two strategies: increasing the prior knowledge of object grasp configuration and reducing the scope of the detection region.

*A. Based on prior knowledge*

Inspired by Faster-RCNN [9], [5], [6], [8] proposed to generate multiple groups of anchors with specific sizes in the image as grasping position priors. According to the confidence score, the model selects the anchor that matches the standard as the candidate and predicts the offset between the candidate anchors and ground-truth to generate the grasp configuration. The method converts the angle data of the grasp rectangle into discrete category data and predicts the angle information by the classification method. [7], [22] proposed to replace the traditional horizontal anchor with an orientation anchor to enrich further the prior knowledge of grasping position. The performance of the method achieves state-of-the-art results at that time.

Using the grasp detection algorithm [5], [6], [8] based on prior knowledge can efficiently indicate potential graspable candidates in the image. Besides, the encoded angle information [7], [12] makes the training and testing of the grasp detection model more efficient compared with the direct regression method. However, these methods do not distinguish between objects and background when detecting grasping positions, which is an indiscriminate global grasp detection. In the whole image, the target only accounts for a tiny part of the image, and most of the rest is the background information that contributes limited to the grasp detection. Therefore, due to the large proportion of background information, indiscriminate grasp detection causes the consumption of computing resources, and there are ambiguous features similar to grasping position in complex background features that affect the accuracy of the grasp detection model.

*B. Region reduction approach*

As opposed to finding grasping positions throughout the scene, [13] proposed a ROIs-based grasp detection model inspired by the target detection algorithm to narrow the scope of grasp detection and distinguish objects in the working scene. The model first detects the objects in the scene and their ROI [9] regions and then uses the features in the ROI region of the target object to predict the grasp configuration rather than the features of the whole scene. Therefore, the model can distinguish different kinds of objects through an object detection algorithm, reduce the scope of grasp detection, and endow the predicted grasping position with object category information so that the model achieves a more intelligent performance. However, due to the axial symmetry of ROI and the irregularity of the target object, it cannot avoid the influence of local information of other objects and background information on the grasp detection model.

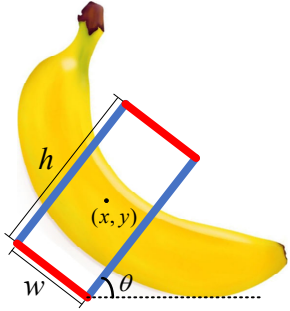

Figure 2. A 5-dimensional grasp representation with location $(x, y)$, rotation $\theta$, gripper opening width $h$ and plate size $w$.

[14] proposed a grasp detection model based on semantic segmentation. The model uses the method of semantic segmentation to determine the pixel regions of different objects and only uses the pixels information in the target object region for grasp detection. This method decreases the interference of background information to the performance of grasp detection. However, the redundant model structure reduces the efficiency of grasp detection.

Inspired by [13] and [14], we propose a dense end-to-end semantic grasp detection model. In the mode, we **reduce the superfluous redundancies structures** in scene perception and propose a **feature attention mechanism** to refine the features areas for grasp detection. The simplified design reduces the training cost and computing resource consumption of the model. Meanwhile, the feature attention mechanism makes the model more sensitive to different objects' structure and texture information and reduces the dependence on background features. Therefore, it is more robust to unstructured environments.

## III. PROBLEM FORMULATION

In this section, we will describe in detail how our model represents the grasp configuration and how does the model predicts grasp configurations of different objects in the multi-object workspace.

### A. Grasp descriptors

Given the features of a target object, the purpose is to identify the target's potential grasp configurations. In this paper, we use the widely used 5-dimensional grasp representation [1], [3], [4], [23], [24] to represent the object grasp configuration. This grasp representation ultimately gives the grasping position of the target object and the opening size and closing orientation of the parallel plate gripper. The 5-dimensional grasp is represented as follow:

$$g = \{x, y, \theta, w, h\} \quad (1)$$

Where $\{x, y\}$ represents the center point of the grasp configuration, that is, the center point of the parallel gripper, $\theta$ is the angle between the closing direction of the parallel gripper and the horizontal direction, $h$ is the opening size, $w$ is the width. An example of grasp representation is shown in Fig. 2.

### B. Grasp detection

As mentioned in Section II, to avoid the model using the global features of the image for grasp detection, we use a semantic segmentation algorithm to distinguish different feature regions of the input image. Thus guidance the model detects grasp configuration in a specific region. In order to train the model's perception of different object features, we labeled the objects in each picture in the dataset with mask labels containing category information.

For the multi-object scene, our model predicts the grasp configurations of each object in the image one by one according to the category of the object. Only the grasp configuration with the highest confidence of each object is reserved for generating the final prediction result of the image.

## IV. METHOD

In this section, we first describe the overall structure of the model, including the feature extractor and feature fusion method. Then we give the details of the object feature attention mechanism and grasp detector. Finally, we introduce the loss function used in the training process of our model.

### A. Architecture

The structure of our model is shown in Fig. 3. It is composed of four parts, backbone unit, feature fusion unit, feature attention unit and grasp detector unit.

The backbone network is a feature extractor that learns the features of different dimensions of input RGB images through five down-sample processes. The lower layers of the feature extractor learn edge information or contour information in the input image, while the higher layers learn abstract semantic information, as shown in Fig. 4. The features learned by different task backbones are comparable, so we can utilize transfer learning to speed up the model's training process. Therefore, in this paper, we choose the pre-trained VGG16 network as the backbone of the model to extract the features of the input image.

The feature fusion unit is used to fuse features of different dimensions generated by the feature extractor. For the deep convolutional neural network, with the depth increasing, the receptive field of the model becomes more extensive, and the resolution of the feature map decreases and contains more abstract information, as shown in Fig. 4. Therefore, for small size objects in the input image, the deep feature map provides little information. In grasp detection and semantic segmentation, shallow features are more suitable for small objects, and depth features are ideal for large objects. We consider the characteristics of these two size feature maps and propose a feature fusion method, which doubles the size of the feature map and concatenates it with the upper feature map so that the final feature map contains rich semantic information and structural information. The feature maps generated after feature fusion are represented by C1, C2, C3, and C4, which are called common feature maps, as shown in Fig. 3. C1 contains abundant features used for semantic segmentation to distinguish pixel regions of different objects in the input image. C2-C4 are used to estimate grasp configurations.

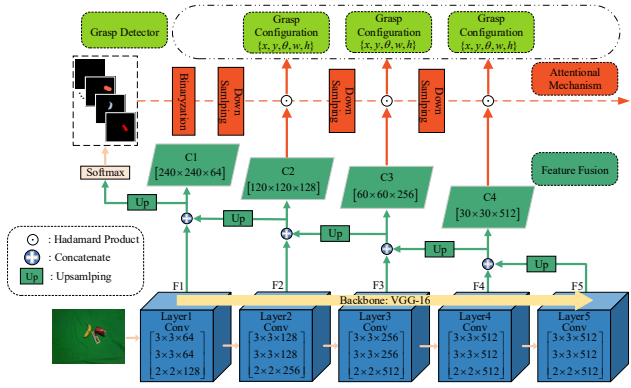

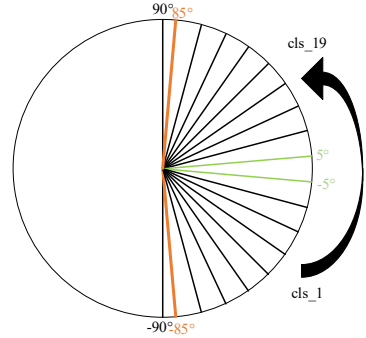

Figure 3. The complete structure of our grasp detection model. This model consists of four components, backbone feature extractor, feature fusion unit, feature attention mechanism, and grasp detector. The binarization module is a feature encoding module based on target position and category.

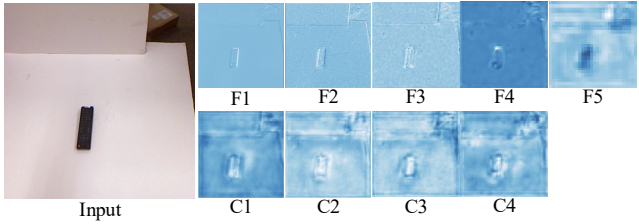

Figure 4. The feature maps are output by each convolution layer. The shallow feature map (F1) mainly contains specific contour features, while the deep feature map (F5) contains more abstract semantic features.

The fused feature maps are finally used to predict grasp configuration as common feature maps. We will explain the feature attention unit and grasp detector unit in detail in the following two parts.

*B. Feature attention unit*

The feature attention unit plays an important role in our work. In this unit, we utilize the C1 feature map to semantic segmentation process based on the multi-classification principle. We divide the result of the semantic segmentation algorithm into multiple feature maps according to object category; each feature map contains only one object's category information and location information. Then, we binarized the feature map generated by the semantic segmentation algorithm. The pixel value of the target object areas is set to 1, and the other background areas are set to 0. After binarization, the feature map is similar to a binarization weight vector. We do the Hadamard Product of the weight vector with the common feature map after down-sampling to highlight the features of the target object area and trowelling the background features, as shown in Fig. 1(a).

The attention mechanism proposed by us enables the model to learn not only the features of grasping position but also the location and semantic features of objects in the training process. In addition, grasping pose is generated under object position and semantics constraints. Therefore, instead of focusing on the features of each pixel in the whole image, our algorithm focuses on the feature areas that are most valuable for predicting the grasp position of the specific object, which surpasses the accuracy of the state-of-the-art grasp detection algorithms.

Figure 5. The mapping relationship between angle values and angle categories. The maximum deviation of this mapping is smaller than 10°.

*C. Grasp Detector*

According to the recommendation of the feature attention mechanism, this module only uses features related to the target to predict grasp configuration. Inspired by [8], we transform the continuous angle data into discrete category data and use the classification method to predict the angle of the grasp rectangle. Eq. 2 describes the mapping relationship between angle data and category data.

$$class\_\theta = round(\frac{\theta + 90}{10}) + 1 \quad (2)$$

This method maps the angle data in the range of -90°-90° into 19 categories, and the category distribution is shown in Fig. 5. At the same time, in order to predict the graspable of different grasping positions, we add another dimension on the basis of 19 categories to predict the score of grasping positions.

*D. Loss function*

We propose a multi-task model, which can realize semantic segmentation and grasp detection simultaneously. Therefore, the model's loss function consists of segmentation loss to supervised segmentation branch training and regression loss and classification loss to supervised grasp detection branch training.

We use the cross-entropy function as the loss function of segmentation. We define the segmentation loss as:

$$L_{Seg\_Loss}(\{p\}) = -\sum_{i=0}^{N-1} p_i \log(\hat{p}_i) \quad (3)$$

Where $N$ is the number of item categories in the dataset. $p = [p_0, ..., p_{N-1}]$ is the one-hot encoding of the sample truth value. When the sample belongs to class $i$, $p_i = 1$; otherwise, $p_i = 0$. $\hat{p} = [\hat{p}_0, ..., \hat{p}_{N-1}]$ is a probability distribution resulting from the model predicted value calculated by softmax and is used to represent the probability that the sample belongs to the $i'th$ category.

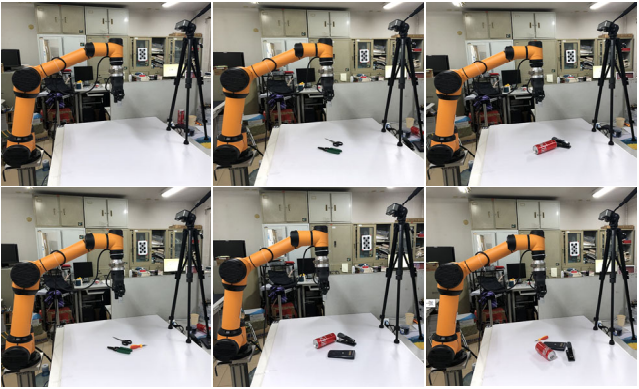

Figure 6. The robot grasping experimental platform consists of an AUBO manipulator and a Surface-120 depth camera placed on a tripod.

We use smooth L1 [9] as regression loss of grasp rectangle prediction. For the multi-object scene, our model predicts the grasp configuration of objects one by one according to the result of semantic segmentation. Therefore, the regression loss in the model training process is composed of the different target loss values. The ground-truth corresponding to different targets will be selected according to the center point's position of the image's annotations. We regard the image annotations contained in the pixel region of the target object as ground-truth; otherwise, it is not. We define the regression loss as follow:

$$L_{reg\_loss}(\{t\}) = \frac{1}{n}\sum_{n=0}^{n-1}\sum_{m\in\{x,y,w,h\}} \text{smooth}_{L1}(t_m - \hat{t}_m) \quad (4)$$

Where $t_m$ is the predicted value of the network, $\hat{t}_m$ is the ground truth of the corresponding grasp configuration. $t$ is a vector representing the four parameters of the grasp rectangle. $n$ represents the number of objects to be grasping in the image.

We use cross-entropy as the classification loss ($L_{cls\_loss}$) to supervise the prediction of angle class and graspable. The classification loss is consistent with the description of $L_{Seg\_Loss}$ and will not be described here. Therefore, we can define the loss function of the grasp detection branch as follows:

$$L_{Grasp\_loss}(\{t\},\{p\}) = L_{reg\_Loss} + L_{cls\_Loss} \quad (5)$$

Finally, the total loss function of our model is defined as:

$$L_{total} = L_{Seg\_Loss} + \alpha L_{Grasp\_loss} \quad (6)$$

In our work, the semantic segmentation branch plays an important role in the model. In order to achieve the best performance of the model, we set $\alpha$ to 0.5.

## V. EXPERIMENTS SET

In this section, we mainly introduce the details of our model training and testing, including the dataset used for model training and testing, the enhancement method of the dataset, the metrics of the model, and the details of the experimental platform.

### A. Dataset

In order to verify the effectiveness of our propose algorithm. The training data and testing data we feed to our network is the Cornell Grasp Dataset. The Cornell Grasp Dataset consists of 885 images containing a single object, and each object has annotated multiple ground-truth grasps. In addition, in order to train the semantic segmentation branch of the model, we annotated the contours and categories of objects in the Cornell Grasp Dataset.

For data preparation, we perform extensive data augmentation to prevent over-fitting during model training. In this paper, we first crop the image to a fixed size of 480*480 according to the center crop, left-crop, and right-crop methods. Then we rotate the image with angle [0, 20, 40, ... ,340] in degree.

### B. Metrics

To demonstrate the performance of our model, we also use a rectangular metric similar to [2], [3], [8] and [11]. This metric considers not only the position, but also the rotation angle, width and height of the predictions. A grasp is considered to meet the metric if it satisfies the following two conditions.

a) The difference between the predicted grasp angle and the ground truth is within $30°$.

b) The Jacquard index of the predicted grasping rectangle and ground-truth is greater than 25%.

The Jacquard index is defined as follow:

$$J(G,\hat{G}) = \frac{G \cap \hat{G}}{G \cup \hat{G}} \quad (7)$$

Where $G$ and $\hat{G}$ are ground truth and predicted grasp rectangle, respectively. $G \cap \hat{G}$ is the overlap area between the predicted grasp rectangle and the ground-truth, and $G \cup \hat{G}$ is the union area between the these two rectangles.

### C. Implementation details

In our work, the backbone network is pre-trained on the ImageNet dataset to speed up model training and prevent overfitting. In the experiment, we fused the output of the Backbone network, C1 feature map after fusion was used for semantic segmentation, and C2-C4 was used for grasp detection experiment, respectively.

Our models are implemented with the PyTorch deep learning framework. For training and testing our model, we use a single NVIDIA GTX2080Ti with 11GB memory. The batch size is set as 1. Our model is trained end-to-end for 100 epochs. We used Adam as the optimizer for model training, and the learning rate was set to 0.0001. We set the learning rate decay of 0.00008 for each epoch.

## VI. RESULTS AND ANALYSIS

In this section, we first describe the details and results of the ablation experiment to prove the effectiveness of our proposed scheme. We then compare our results with other grasp detection algorithms. Finally, we verify the performance of our model in single-object and multi-object

TABLE I. SELF-COMPARISON EXPERIMENTS ON CORNELL GRASP DATASET UNDER DIFFERENT CONDITIONS

| No. | Method | Setting | Accuracy (%) |
|---|---|---|---|
| 1 | Baseline | RGB, F3 | 94.5 |
| 2 | Baseline, Feature attention | RGB, F3 | 95.41 |
| 3 | Baseline, Feature fusion | RGB, C3 | 97.15 |
| 4 | Baseline, Feature attention, Feature fusion | RGB, C4 | 95.76 |
|   |   | RGB, C3 | **98.38** |

TABLE II. ACCURACY OF DIFFERENT METHODS

| Author | Backbone | Input | Accuracy (%) |
|---|---|---|---|
| Guo et al. [5] | - | RGB | 93.2 |
| Zhang H. et al. [13] | ResNet-101 | RGB | 93.6 |
| Fu-Jen. et al. [6] | ResNet-50 | RGB | 96.0 |
| Shao Z. et al. [2] | - | RGB-D | 95.51 |
| Ours | VGG-16 | RGB | **98.38** |

scenes in the real world through physical grasping experiments.

*A. Ablation Study*

In this part, we demonstrate the contribution of each component to the model performance through a series of self-comparison experiments. Table I shows the detailed comparative experimental results. In the first two experiments, the model uses the F3 features to predict grasp configuration, and in the second experiment, we introduce a feature attention unit. The experimental results show that the model's performance is slightly improved when RGB images are used as input. The reason is that the background of the center-cropped image contains very few ambiguity features similar to the object grasping position, as shown in Fig. 4. Therefore, the model can easily determine the grasping position of the object by using the F3 feature map. We use the first and third sets of experiments to explore the effectiveness of the feature fusion mechanism on model performance. We can see that the model's accuracy with the feature fusion mechanism outperforms the Baseline by 2.65% when RGB images are used as input. This is because the C3 features after feature fusion contain both detailed information and semantic information. As described in Fig. 4, object information in the C3 features is more affluent than that in the F3 features. At the same time, by comparing the experimental results of No. 2 and No. 3, we can see that the feature fusion mechanism has a more significant contribution to grasp detection in the case of a simple background. We believe that this result will improve when the environment becomes complex. In the last experiment, the model included feature fusion and feature attention mechanism, and the model performance was 3.88% higher than the Baseline when RGB images were input into the model. This demonstrates that the method and contribution we proposed effectively improve the performance of the grasping detection model.

We only used C3 and F3 features to predict the grasp configuration during the entire ablation experiment. The input image size in our model is 480*480, and the object occupies only a tiny part of pixels in the image. After 16 times down-sampling, there are few features of the object left, as shown in Fig.4. F5. Moreover, F5 features almost contain no structural information but more semantic information, which is not reasonable for grasp detection models. In addition, it is difficult to filter the diminished and distorted features. In contrast, C3 and F3 features are more suitable for grasp detection because C3 and C4 provide more delicate features containing more detailed image information with little loss of network depth.

*B. Results for Single-object Grasp*

Table II summarizes the results of the previous methods and our proposed method. We can see that our algorithms achieve the highest accuracy. Compared with the method of using anchor with position prior submitted by Guo et al. [5], our method achieves a 5.18% gain of the final performance. This demonstrates that the feature attention mechanism proposed by us can effectively guide the model to focus on the features of the object itself and minimize the influence of irrelevant features on the model performance. Compared with the ROI-based grasp detection algorithm proposed by Zhang et al. [13], our algorithm using lightweight Backbone has achieved a 4.78% improvement in performance. The reason may be that the more detailed feature extraction method of our algorithm provides excellent aid to the improvement of model performance.

Table III summarizes the results of our method and the state-of-the-art methods on the single-object Cornell Grasp Dataset. In the case of the VGG-16 backbone network, the comparison with Zhang et al. [25] shows that our approach improves the performance of the grasp detection model. At the same time, compared with the state-of-the-art method, our model's accuracy has the lowest variability under different Jacquard thresholds. The reason may be that our model learns and predicts grasp configurations from the overall features of the object rather than the local features of the object. Therefore, the model is not disturbed by local features, and it is easier for the model to find grasping positions that match human grasping habits.

Table IV summarizes the accuracy of our algorithm and Zhang's algorithm [25] under different angle thresholds. We can see that our algorithm can still maintain satisfactory performance even when the angle threshold is under the harsh condition of 10°. At the same time, our model has good robustness to the change of angle threshold from the perspective of accuracy variation amplitude. This also proves that the object-centric grasping detection method can obtain better performance.

In Fig. 7, we visualized semantic grasping results and the corresponding ground-truth of our model in the test set of the Cornell Grasp Dataset. Our model uses the results of semantic segmentation to filter the features entering the grasp detector, so the semantic segmentation branch provides semantic information for the grasp detector, making the generated grasp configuration with semantic information. As shown in Fig.7, the first and third rows are the predicted results of the model, containing the grasp configuration and corresponding semantic information, confidence score. The second and fourth rows are ground truth.

TABLE III  ACCURACY UNDER DIFFERENT JACCARD THRESHOLDS

| Author | Backbone | Input | Accuracy (%) | | | | Variability (%) |
|---|---|---|---|---|---|---|---|
| | | | 20% | 25% | 30% | 35% | |
| Guo et al. [5] | - | RGD | 93.8 | 93.2 | 91.0 | 85.3 | 9 |
| Chu et al. [6] | ResNet-50 | RGD | - | 96.0 | 94.9 | 92.1 | 4 |
| Zhang et al. [25] | VGG-16 | RGD | 98.0 | 97.5 | 93.8 | 87.5 | 10.7 |
| | ResNet-101 | RGD | 98.88 | 98.88 | 96.4 | 93.7 | 5.2 |
| **Ours** | VGG-16 | RGB | **99.46** | **98.38** | **98.03** | **96.39** | **3.1** |

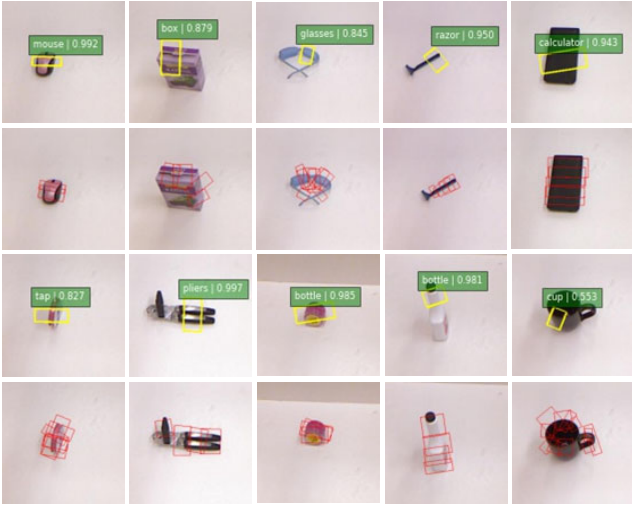

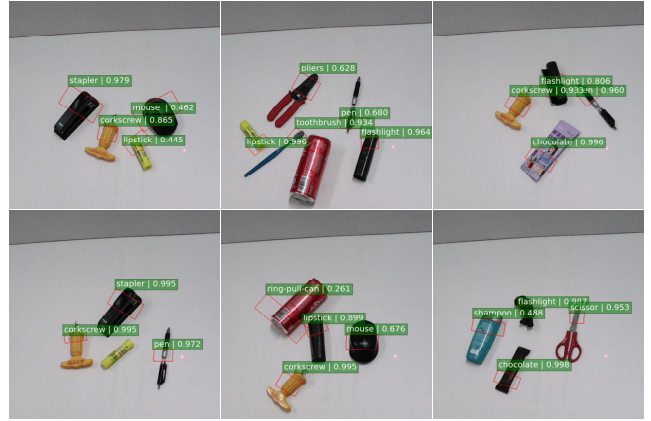

Figure 7. The results of semantic grasp detection. The first and third rows are the predicted results of the model, containing the grasp configuration and corresponding semantic information. The second and fourth rows are ground truth.

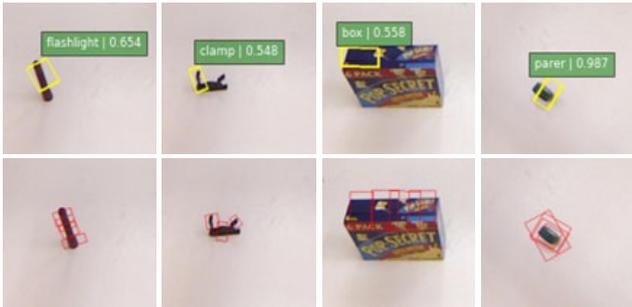

Figure 8. Unsuccessful grasp detection from our model. The first row is the detection result. The second row is the corresponding ground truth.

Some detection results judged to be incorrect by the model are shown in Fig. 8. We can see that the grasping positions predicted by the model are all correct, but the predicted angles are biased. This is the main reason to be judged as an unsuccessful grasp. On the other hand, even though there is a significant deviation between the predicted grasp configuration and the ground-truth in angle, there are still some predicted grasp configurations that the robotic can successfully grasping in practice, such as the situation in the second and fourth image. Therefore, incomplete annotated datasets are also a factor that affects the performance of the model.

Figure 9. Results of multi-object grasp detection. Our model was trained on the Cornell single-object dataset but is well suited for detecting grasp configurations of unseen objects in more complex environments.

TABLE IV  
ACCURACY UNDER DIFFERENT ANGLE THRESHOLDS

| Author | Backbone | Input | Accuracy (%) | | |
|---|---|---|---|---|---|
| | | | 10° | 20° | 30° |
| Zhang et al [25] | VGG-16 | RGD | 79.0 | 95.1 | 97.5 |
| Ours | VGG-16 | RGB | 91.93 | 97.24 | **98.38** |

*C. Results for Multi-object Grasp*

In this section, we apply the proposed model to a multi-object grasp detection task, using ordinary objects that do not appear in Cornell Grasp Dataset. The model was only trained on the Cornell Grasp Dataset, and each image in the dataset contained only a single object with a simple scene. The object feature attention mechanism proposed by us guides the model to use the features of the object itself to generate the grasping position. Therefore, the model can be more easily migrated to the multi-target complex environment.

TABLE V PHYSICAL GRASPING ACCURACY

| Object | Detected | | Physical | | mIoU |
|---|---|---|---|---|---|
| | Single | Multiple | Single | Multiple | |
| stapler | 10/10 | 10/10 | 10/10 | 10/10 | 74.98 |
| cup | 10/10 | 9/10 | 8/10 | 8/10 | 75.52 |
| mouse | 10/10 | 9/10 | 8/10 | 7/10 | 77.82 |
| ring-pull-can | 10/10 | 8/10 | 7/10 | 8/10 | 76.72 |
| toothbrush | 9/10 | 7/10 | 8/10 | 7/10 | 72.42 |
| corkscrew | 10/10 | 9/10 | 10/10 | 9/10 | 72.53 |
| lipstick | 10/10 | 8/10 | 10/10 | 8/10 | 81.53 |
| screwdriver | 8/10 | 9/10 | 7/10 | 9/10 | 73.8 |
| tape | 10/10 | 8/10 | 8/10 | 7/10 | 75.05 |
| spectacle-case | 9/10 | 10/10 | 9/10 | 9/10 | 75.78 |
| average (%) | 96% | 87% | 85% | 82% | 75.62 |

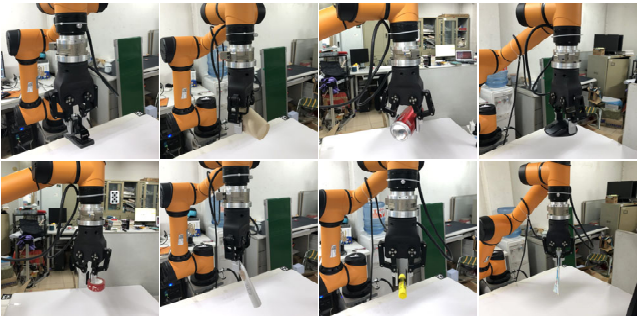

Figure 10. Physical Grasping. The pictures of a robot performing a grasping.

Fig. 9 are the results of our model in a multi-object environment. The categories of the tested objects are not completely included in the Cornell Grasp Dataset, so there will be a small number of object category recognition errors. It can be seen from the recognition results that our model successfully locates the grasping position of objects and predicts their categories in a multiple new object scene with very few false positives.

### D. Physical Grasping

**Experimental Platform** In our physical grasping experiment, we use an AUBO manipulator with 6-Dof made in China as the executor. The operating platform is in front of the manipulator, and its surface and the base of the manipulator are in the same horizontal plane. We use a Surface-120 camera to collect RGB-D images of the working scene in the physical experiment. The camera is mounted on the opposite side of the manipulator using a tripod about 60cm higher than the working plane with a depression angle of 80°. The parallel plate gripper we used was not equipped with a force sensor, so in order to successfully grasping different types of targets, we set the closing position of the gripper before the grasping experiment. The complete experimental platform is shown in Fig. 6.

**Evaluation Strategy** To evaluate our algorithm's performance and generalization ability in the real world, we test the success rate of grasping a specific object in single-object and multi-object scenarios. In the single object experiment, we put an object alone in different positions in the working scene of the manipulator with different postures to verify the grasp detection accuracy of the model for an object in different situations. In the multi-object experiment scene, objects with different types and shapes are randomly placed on the table. Compared with the single-object scene, the multi-object scene is more complex and contains more disturbance. Hence, in this experiment, we focus on the manipulator to grasping the specific object in the multi-object scene to verify the generalization ability of our model.

The objects of our physical experiments include stapler, cup, mouse, ring-pull-can, toothbrush, corkscrew, lipstick, screwdriver, tape, and spectacle-case. In the physical experiment, we define the symbol of successful grasping as the manipulator successfully grasping the object and stably leaving the table. Based on the above conditions, we try ten grasping experiments for each of the above objects under different conditions and record the success rate.

**Results** The results of the physical grasping experiment are shown in Table V and Fig 10. None of the objects used in the experiment appeared in Cornell Grasp Dataset to verify our model's generalization ability. From Table 5, we can see that our model achieves a success rate of 96% and 85% for prediction and physical experiments, respectively, under single-object conditions. And in the multi-object scenarios, our model achieves 87% and 82% success rates for prediction and physical experiments, respectively. Meanwhile, our model's mean intersection-over-union of different object segmentation is 75.63. These experimental results demonstrate that our model can not only be applied to single object scenes but also be extended to complex multi-object scenes to grasp the specific target.

## VII. Conclusion and Future Work

In this paper, we propose a novel semantic grasp detection model. Inspired by human grasping habits, we propose a feature attention mechanism that guides the model to focus on the features of the object itself, which effectively improves the performance of grasp detection. Moreover, based on this mechanism, our algorithm realizes object-centric semantic grasp detection. Besides, the feature fusion mechanism is used for robotic grasping detection, which enriches the semantic and fine-grained information of object features. The results of ablation experiments also demonstrate the effectiveness of our proposed methods. Finally, we demonstrate that our model can be applied to both single-object and complex multi-object scenarios through physical grasping experiments.

In the future, we will focus on the research of grasp detection and grasp strategy for stacked objects. We will add a multi-scale feature learning branch to the existing feature map and fine-tune all the sub-branches in the model. Multi-scale feature learning branches are used to predict the relative position of objects in the robot workspace. Then, we can make a scientific grasping sequence according to the relative position of objects to realize the grasping of specific objects in the stacking scene. Moreover, as our model can detect the grasp configuration in real-time, accurate prediction of spatial position relationship of objects can provide position constraints for the robot during execution.